\documentclass[letterpaper, 10 pt, conference]{ieeeconf}  % Comment this line out if you need a4paper

\IEEEoverridecommandlockouts                              % This command is only needed if 
\usepackage{multirow, makecell}
\usepackage{hhline}
\usepackage{array, booktabs}

\usepackage{graphics} % for pdf, bitmapped graphics files
\usepackage{epsfig} % for postscript graphics files
\usepackage{amsmath} % assumes amsmath package installed
\usepackage{amssymb}  % assumes amsmath package installed
\usepackage{bm}
\usepackage{multirow, makecell}
\title{\LARGE \bf
Multi-modal Probabilistic Prediction of Interactive Behavior via an Interpretable Model
}

\author{Yeping Hu, Wei Zhan, Liting Sun and Masayoshi Tomizuka % <-this % stops a space\
	\thanks{*This work was supported by Momenta.}
	\thanks{Y. Hu, W. Zhan, L. Sun, and M. Tomizuka are with the Department of Mechanical Engineering, University of California, Berkeley, CA 94720 USA {\tt {[yeping\_hu, wzhan, litingsun, tomizuka@berkeley.edu]}}}
}

\makeatletter
\def\endthebibliography{%
	\def\@noitemerr{\@latex@warning{Empty `thebibliography' environment}}%
	\endlist
}
\makeatother
\begin{document}

\maketitle
\thispagestyle{empty}
\pagestyle{empty}

%%%%%%%%%%%%%%%%%%%%%%%%%%%%%%%%%%%%%%%%%%%%%%%%%%%%%%%%%%%%%%%%%%%%%%%%%%%%%%%%
\begin{abstract}
For autonomous agents to successfully operate in real world, the ability to anticipate future motions of surrounding entities in the scene can greatly enhance their safety levels since potentially dangerous situations could be avoided in advance. While impressive results have been shown on predicting each agent's behavior independently, we argue that it is not valid to consider road entities individually since transitions of vehicle states are highly coupled. Moreover, as the predicted horizon becomes longer, modeling prediction uncertainties and multi-modal distributions over future sequences will turn into a more challenging task. In this paper, we address this challenge by presenting a multi-modal probabilistic prediction approach. The proposed method is based on a generative model and is capable of jointly predicting sequential motions of each pair of interacting agents. Most importantly, our model is interpretable, which can explain the underneath logic as well as obtain more reliability to use in real applications. A complicate real-world roundabout scenario is utilized to implement and examine the proposed method.

%Real-world scenarios demand a model of uncertainty of such predictions, as predictions become increasingly uncertain – in particular on long time horizons. While impressive results have been shown on point estimates, scenarios that induce multi-modal distributions over future sequences remain challenging.

\end{abstract}

%%%%%%%%%%%%%%%%%%%%%%%%%%%%%%%%%%%%%%%%%%%%%%%%%%%%%%%%%%%%%%%%%%%%%%%%%%%%%%%%
\section{Introduction}
The idea of predicting the future behavior of statistical time series has a wide range of application in economics, weather forecast, intelligent agent systems, etc. The autonomous vehicle is one of the well-known intelligent agents and it is expected to predict behaviors of other road entities. Accurate and reasonable prediction is a prerequisite of performing reliable tasks involving motion planning, decision making, and control.

There have been numerous researchers working on behavior prediction problems and many approaches have been explored to solve such problem in the autonomous driving area. Some of these approaches only performed point estimate by assuming that the environment is deterministic. However, they failed to taking into account the uncertainty of future outcomes caused by partial observation or stochastic dynamics, which will induce the lost of information that capture the real physical interactions. Therefore, in this paper, we take into account the uncertainty of drivers as well as the evolution of the traffic situations and try to predict possible behaviors of multiple traffic participants several steps into the future.

\begin{figure}[htbp]
	\centering
	\includegraphics[scale=0.53]{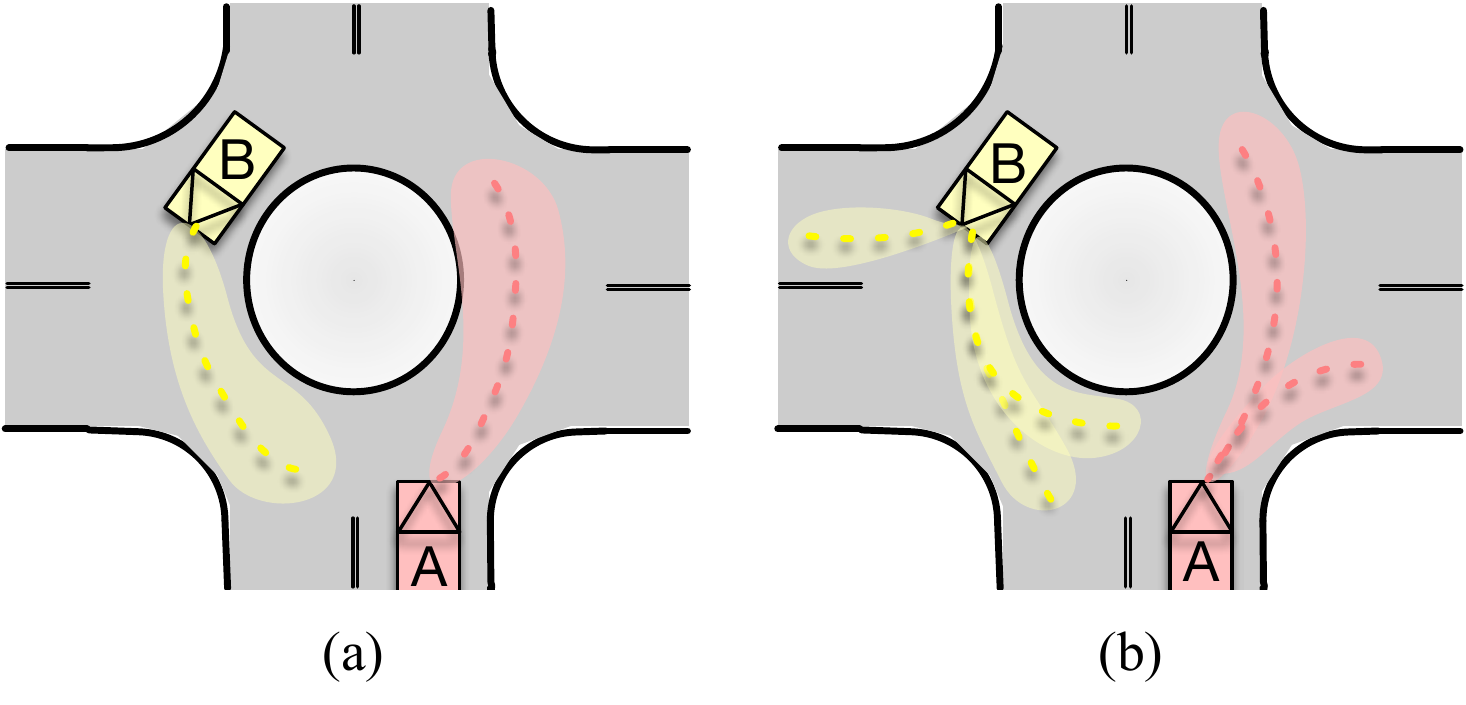}
	\caption{A demonstration of (a) single-modal and (b) multi-modal predicted distributions.}
	\label{fig:first_image}
\end{figure}

When dealing with uncertainties for sequential prediction problems, two aspects need to be further discussed: the estimation of \textit{multi-modal} output distribution and the \textit{interpretability} of the output samples. The multi-modal property of a method can be regarded as having different motion patterns in the outputs, which is illustrated in Fig.~\ref{fig:first_image}. As was reviewed and summarized in \cite{wei_zhan}, motion patterns can be categorized hierarchically into route, pass-yield and subtle patterns in various kinds of scenarios. For routing information, motion patterns can be regarded as discrete. In contrast, predicted future trajectories are expected to have continuous motion patterns, where the agent's speed could increase, decrease or change randomly for a sequence of future motions. 

The model interpretability is also a crucial aspect that needs to be considered. Since the output uncertainty is usually achieved by sampling data points from some learned distributions, it is necessary to reason about what causes the motion pattern to vary among samples. However, most of the commonly used approaches cannot provide much insight on the structure of the function that is being approximated, especially for learning-based methods.

%However, in terms of interactions between two agents, there will be various possible joint motions and thus the motion patters should be treated as continuous. Also, the output uncertainty is usually achieved by sampling data points from some learned distributions or from manually added random noise \cite{xx}. As we want to output a sequence of future motions, the agent's speed could increase, decrease or change randomly in the next few time steps. Therefore, the sampled trajectories should represent different motion patterns and thus it is necessary to reason about what causes the pattern variation of each sample. However, most of the commonly used approaches cannot provide any insight on the structure of the function being approximated, especially for learning-based methods, which are untrustable to use in real life. % and it is hard to have any explanation. 

The contributions of this paper are in four folds: First, we proposed a multi-modal probabilistic prediction structure for autonomous vehicles using only a single learning-based model. Second, we considered the sequential motion prediction of each pair of interacting vehicles. Third, the proposed model is interpretable as we are able to explain each sampled data point and relate it to the underlying motion pattern. Last but not least, we trained and tested our method under a complicated roundabout scenario, which adds more difficulties to the behavior prediction problem.

The remainder of the paper is organized as follows: Section II provides a brief overview of works related to interpretable models, trajectory prediction, and multi-modality. Section III provides the detailed explanation of the proposed approach; Section IV discusses an exemplar scenario to apply our method; evaluations and results are provided in Section V; and Section VI concludes the paper.

%\begin{itemize}
%	\item We used $\beta$-CVAE
%	\item We only use one model instead of multiple models to generate the multi-modal prediction distribution
%	\item Our model is more interpretable and we can even discover the underlying mechanism of how samples are generated 
%	\item We add the additional intention indicator to the input features (Include pictures or algorithms)
%	\item It will learn multimodal interactive behavior for each intention
%	\item We did not assume one of the vehicle has to keep its lane, the model is able to predict weaving trajectories
%	\item Our method is able to apply to any vehicle pairs in the scene
%	\item Can be regarded as unsupervised learning since the $z$ space is learned by itself.
%\end{itemize}

\section{Related Works}
%\noindent\textbf{Structured Representation}.\\
\noindent\textbf{Interpretable Models}\\ 
\-\hspace{0.15cm} To solve prediction problems for autonomous vehicles, many researchers utilized traditional methods such as constant velocity (CV), constant acceleration (CA), Intelligent Driver Model (IDM), and Kalman Filter (KF) \cite{PF}. However, these methods work well only under simple driving scenarios and their performances degrade for long-term prediction as they ignores surrounding context. As these models can easily fail when scaled to complicated traffic scenes, classical machine learning models are usually used instead, such as Hidden Markov Model (HMM) \cite{HMM_2}, Bayesian Networks (BN) \cite{BN}, Gaussian Process (GP) \cite{GP}, and Inverse Reinforcement Learning (IRL) \cite{IRL}\cite{IRL_2}. The aforementioned methods either utilize some known transitions models or incorporate hand-designed domain knowledges to enhance the prediction performance. Although these works consist of interpretable methodology, their performances are usually worse than most learning-base methods that lack interpretations. 

The success of deep learning in many real-life applications motivates research on its use for motion prediction and related methods include Mixture Density Networks (MDN) \cite{Yeping_MDN}, Recurrent Neural Networks (RNN) \cite{RNN}, and Convolutional Neural Networks (CNN) \cite{traj_CNN}. Deep learning models can achieve high accuracy but at the expense of high abstraction which cannot be trusted.

Recently, numerous researchers tried to reason about learning-based methods by utilizing the idea of variational autoencoder (VAE) \cite{VAE_original} which is a latent variable model that is able to learn a factored, low-dimensional representation of data. \cite{VAE_interpret1} developed a framework for incorporating structured graphical models in the encoders of VAE that allows them to induce interpretable representations through approximate variational inference. \cite{VAE_interpret2} proposed a novel factorized hierarchical VAE to learn disentangled and interpretable latent representations from sequential data.  Our goal is to develop an interpretable architecture for behavior prediction based on the latent variable model such that features involved in modeling can be described through latent codes and explainable future motions can be generated.

%CVAE has the similar structure as VAE but its main purpose is for prediction instead of data/image generation

%\cite{VAE_roundabout} describe features of multi-vehicle trajectories through latent codes

%Interpretable models include: models that assume the output distribution assembles some know distribution function such as Gaussian. 

%Kingma and Welling \cite{xx} introduced the variational autoencoder (VAE) as a latent variable model for efficient maximum marginal likelihood learning and to learn a factored, low-dimensional representation of data.
% the latent space in VAE such that features involved in modeling can be described through latent codes.

%\begin{itemize}
%	\item \cite{VAE_1} Modeling future distributions over the entire traffic scene is a challenging task, given the high dimensional feature space and complex dynamics of the environment. Hence many approaches simply the problem by assuming that the environment is deterministic, where only one future outcome is possible. However, real-world scenario usually have stochastic dynamics and by assuming a deterministic model, information that capture the real physical interactions can be lost. 
%\end{itemize}

\noindent\textbf{Trajectory Prediction}\\
%Note that MDN is also able to generate probabilistic prediction, but it suffers from computational cost if it wants to directly predict a trajectory sequence. MDN is okay but MDNs are often difficult to train in practice due to numerical instabilities when operating in high-dimensional spaces.
\-\hspace{0.15cm} There are variety of works dealing with trajectory prediction problems for road entities such as vehicles and pedestrians. \cite{traj_LSTM1} proposed a Long Short-Term Memory (LSTM) encoder-decoder model to predict a multi-modal predictive distribution over future trajectories based on maneuver classes. \cite{traj_HMM1} applied the Hidden Markov Model (HMM) to predict the trajectories for individual driver. \cite{traj_image1} combined CNN and LSTM to predict multi-modal trajectories for an agent on a bird-view image. The main limitation of these works, however, is that they only predict motions for one selected agent without considering the influence of other agents with potential interactions. 

Since the motion of an agent can be largely influenced by other surrounding agents, some researchers began to tackle the scene prediction problem. Modeling future distributions over the entire traffic scene is a challenging task, given the high dimensional feature space and complex dynamics of the environment. \cite{traj_LSTM2} and \cite{traj_LSTM3}  brought forward a LSTM-based structure to predict the most possible $K$ trajectory candidates for every vehicles over occupancy grid map. \cite{traj_DBN1} utilized the Dynamic Bayesian network (DBN) for behavior and trajectory prediction. In \cite{Yeping_CVAE}, the authors proposed a hierarchical scene prediction framework, where the Conditional Variational Autoencoder (CVAE) was used in their lower module to predict continuous motions for multiple interacting road participants.

%Feedforward network with Gaussian Mixture \cite{Yeping_MDN}\cite{MDN_1} is also capable of generating probabilistic prediction but it is often difficult to train in practice due to numerical instabilities when operating in high-dimensional spaces such as predicting a motion sequence.
\noindent\textbf{Multi-Modality}\\
\-\hspace{0.15cm} There exists a number of studies addressing the problem of modeling multi-modality. Feedforward network with Gaussian Mixture \cite{Yeping_MDN}\cite{RNN} is usually applied to solve multi-modal regression tasks but it is often difficult to train in practice due to numerical instabilities when operating in high-dimensional spaces such as predicting future sequences.

Other works solved this problem by utilizing different regression models for different possible discrete intentions of road entities \cite{GP}\cite{Yeping_CVAE}. However, when the intention space is large or data is insufficient, such method becomes inefficient and the model will even suffer from over-fitting problems. Alternatively, \cite{traj_CNN}\cite{VAE_1} treated the discrete intention as one of the state input to the proposed structure to generate different types of output.

\section{Approach}
In this section, we first introduce the main algorithm of the proposed behavior prediction approach. Then the details of the intention prediction method are illustrated. 
%\subsection{Conditional Variational Autoencoder(CVAE)}
\subsection{Interactive Behavior Prediction}

Our proposed method is based on the structure of CVAE which has a similar encoder-decoder structure as the typical VAE. 
%The latent space of CVAE and VAE are interpretable, which is able to ......
Two types of conditional input are considered in our model structure: historical scene information and inferred driving intention. 

We focus on predicting human drivers' interactive behaviors between two vehicles: vehicle $A$, (denoted by $(\cdot)^A$), and vehicle $B$, (denoted by $(\cdot)^B$). Both vehicles are regarded as the predicted vehicle and we are interested in jointly predicting their behaviors, while taking into account any internal correlations.
Note that it is trivial to convert the output joint distribution to a conditional distribution by treating one of the predicted vehicles as the ego vehicle and obtain the behavior prediction of the other. However, we will not address further on such problem setting in this paper.

For a given vehicle, we use $\xi$ to represent its actual trajectory and $\hat{\xi}$ as the trajectory we predict. At the current time step $t$, we denote the vehicle's historical trajectory as $\xi_{(t-T_1):t}$ and its future trajectory as ${\xi}_{t:(t+T_2)}$, where $T_1$ and $T_2$ represent the number of time steps into the past and future, respectively. Moreover, we denote $\mathcal{I}$ as the discrete intention of a vehicle and $\mathcal{E}$ as the environment information that contains states of surrounding vehicles.
\begin{figure*}[htbp]
	\centering
	\includegraphics[scale=0.45]{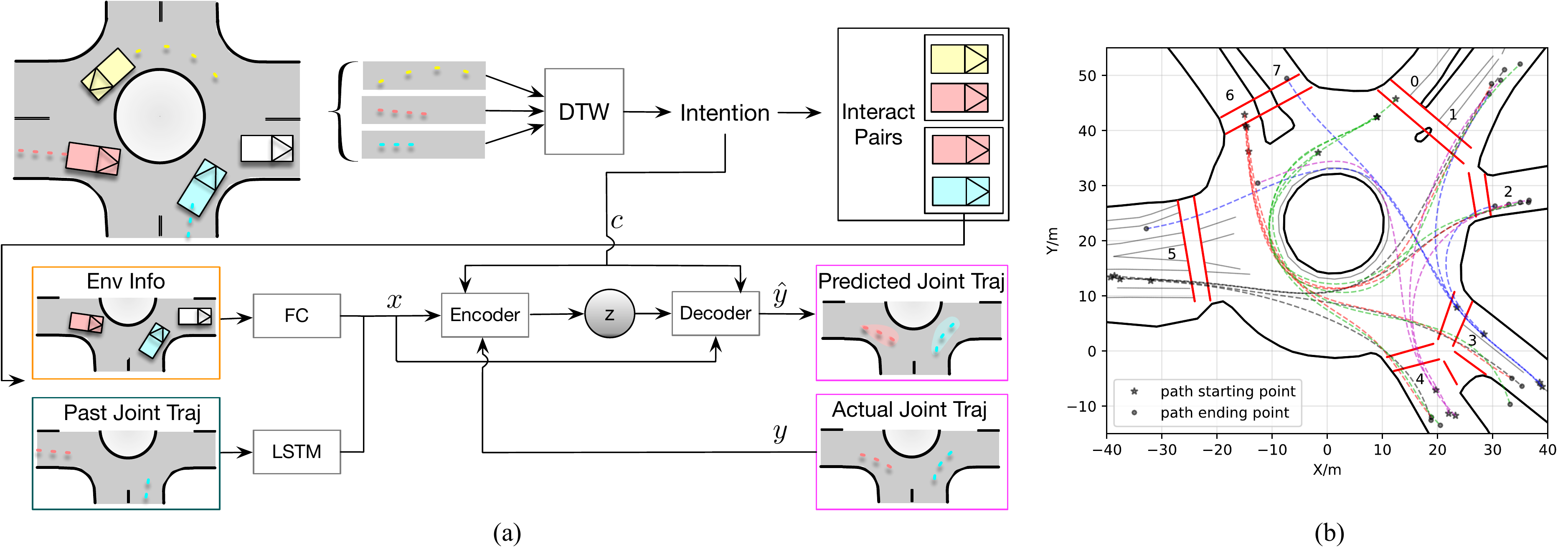}
	\caption{(a) The overall structure of the proposed method. (b) Roundabout map and all reference path.}
	\label{fig:full_structure}
\end{figure*}

Given the historical trajectory and driving intention of two interactive vehicles, along with the environment information, the objective of estimating probabilistic joint trajectories can be expressed as:
\begin{equation}
P(\bm{\xi}^{\{A,B\}}_{t:(t+T_2)} \mid \bm{\xi}^{\{A,B\}}_{(t-T_1):t}, \bm{\mathcal{I}}^{\{A,B\}}, \mathcal{E}).
\end{equation}
To formulate the problem in the CVAE structure, we let the encoder, $q_\theta$, take the input $x$ as a learned embedded space of historical trajectory and environment information, $c$ as the intention vector $\bm{\mathcal{I}}^{\{A,B\}}$, and $y$ as the actual trajectory $\bm{\xi}^{\{A,B\}}_{t:(t+T_2)}$, to ``encode" them into a latent $z$-space. Then the decoder, $p_\phi$, takes $x$ and $c$ as input, to ``decode" the sampled values from $z$-space back to the output $\hat{y}$, which corresponds to the predicted future trajectories $\hat{\bm{\xi}}^{\{A,B\}}_{t:(t+T_2)}$. To enable backpropagation, the network is trained using the reparameterization trick \cite{VAE_original} such that the latent variables can be expressed as: 
\begin{equation}
	z = \mu_\theta(x,c,y) + \sigma_\theta(x,c,y) \times \epsilon, \quad \epsilon \sim \mathcal{N}(0, I).
\end{equation}

Here, $\theta$ and $\phi$ are the parameters of the encoder and decoder network, respectively. To learn these parameters, we can optimize the variational lower bound:
\begin{equation}
\begin{split}
	\mathcal{L} = -\mathbb{E}_{q_\theta(z|x,c,y)}&\big[\log p_\phi(y|x,c,z)\big] \\ & + \beta D_{KL}(q_\theta(z|x,c,y)||p(z)),
\end{split}
\end{equation}
where the model is forced to perform a trade-off between a good estimation of data log-likelihood and the $KL$ divergence of the approximated posterior $q_\theta$ from prior $p(z)$ which, in our case, is assumed as unit Gaussian $\mathcal{N}(0,I)$. We also utilize the hyperparameter $\beta$ to control the training balance between the two losses for better performance.
%\begin{equation}
%\begin{split}
%\mathcal{L} &= ||\bm{\xi}^{\{A,B\}}_{t:(t+T_2)} - \hat{\bm{\xi}}^{\{A,B\}}_{t:(t+T_2)}||^2 + \beta D_{KL}\Big[Q(z \mid \bm{\xi}^{\{A,B\}}_{(t-T_1):t} \\ & ,\bm{\xi}^{\{A,B\}}_{t:(t+T_2)}, \bm{\mathcal{I}}^{\{A,B\}}) - P(z \mid \bm{\xi}^{\{A,B\}}_{(t-T_1):t} )\Big]
%\end{split}
%\end{equation}

At test time, we can directly sample from $\mathcal{N}(0,I)$ as the latent variable input and only use the decoder to obtain the predicted joint distribution. 
%Therefore, at test time, we can sample from the distribution $P(\hat{\bm{\xi}}^{\{A,B\}}_{t:(t+T_2)} \mid \bm{\xi}^{\{A,B\}}_{(t-T_1):t}, \bm{\mathcal{I}}^{\{A,B\}})$ by sampling directly from $\mathcal{N}(0, I)$. 
%Normally, this is the end of the story for most approaches using similar structures, but we want to be one step further and try to explore the interpretability underneath the proposed model. 
Notice that among the three input of the decoder network, only $x$ is fixed at a given time step while both $c$ and $z$ are non-deterministic factors. Therefore, in the following section, we will analyze the effects of these factors to the output trajectories, demonstrate how the proposed model can estimate multi-modal distributions over future sequences, and explore the interpretability underneath the model.

%besides the fixed input $x$, the final result also depends on the intention factor $c$ of two predicted vehicles. Therefore, if we fix the input historical trajectories and sampled $z$ vector, by changing the intentions $\bm{\mathcal{I}}^{\{A,B\}}$, we are expected to have different sampled future trajectories that could reasonably interpret the corresponding intentions. Instead, if we only vary the $z$ vector along each of its dimension, we are expected to generate predicted joint trajectories with distinct features even under a fixed intention. We will explain more on 

\subsection{Intention Prediction}
\subsubsection{Bayesian Approach}
During the scene evolution, we use a Bayesian approach to predict each vehicle's intention $i$, based on history observation $h$. In this problem, we consider the vehicle's past trajectory as the observation $h$ since an agent's potential intent can largely influence its trajectory. Therefore, the Bayesian equation can be written as:
\begin{equation}
	f(i|h) = \frac{f(h|i)f(i)}{\sum_{i}f(h|i)f(i)},
\end{equation}
where $f(\cdot)$ represents the probability density function. The term $f(i)$ is the prior belief of the intention and is initialized with a known distribution according to initial observation; $f(h|i)$ is the likelihood of observing $h$ for a given intention $i$; and the denominator is a normalization term. 

%assume that the intention of an agent can be inferred from the historical trajectory and thus $h$ in our case is the past trajectory of the vehicle.
\subsubsection{Dynamic Time Wraping (DTW)}
%\cite{DTW_origin}\cite{traj_HMM1}\cite{qi2018intent}\cite{traj_metric}\\
The dynamic time warping (DTW) distance as proposed in \cite{DTW_origin} is a trajectory measure that can be used on general time series. DTW does not require both trajectories to have the same length. Instead, DTW measures the temporal changes that are necessary in order to warp one trajectory into another.  

If we consider driving intention as pursuing some goal location such as one of the exit branch in roundabout scenario or left/right turn in intersections, we are able to obtain a reference driving path for each intention. Therefore, we can use the DTW algorithm to determine the likelihood of an observed trajectory $h$ given a reference path $r_i$ assuming the agent has intention $i$:

\begin{equation}
	f(h|i) = \frac{e^{-D(r_i,h)}}{\sum_{i}{e^{-D(r_i,h)}}},
\end{equation}
where $D(r_i, h)$ is the cost calculated by the DTW algorithm. The smaller the cost is, the closer the observed trajectory is to the reference path, and thus the higher the probability is for intention $i$. In fact, we are interested in the DTW value between the observed trajectory and a segment of the reference path closer to the trajectory instead of the full reference path.

\subsubsection{Select Interacting Pairs}
After obtaining the intention probabilities of each vehicle in the scene at a given time step, we can determine whether any of the two vehicles have potential interaction according to their corresponding reference path. If all potential reference path of two vehicles have no crossing point, the vehicles' future trajectories will be independent from each other and thus no attempt is needed to further predict their joint motions. In this work, we make an assumption that the interaction happens only between two agents but there can be multiple interacting pairs in the scene concurrently.

\subsubsection{Avoid Constantly-Changed Prediction Results}
To avoid rapid fluctuation of the likelihood distribution, we perform the aforementioned intention prediction algorithm for at least every other 0.4s. Here we assume that a certain driving intention will not change within a short period of time especially for the roundabout scenario where the driver already know his/her intended road branch to exit.

\section{Experiments}
In this section, we use an exemplar roundabout scenario
to apply our proposed behavior prediction method. The
data source and details of the problem formulation are presented.
\subsection{Real Driving Scenario}
\subsubsection{Dataset}
The driving data we used was collected by our Mechanical Systems Control Lab at a single-lane roundabout in Berkeley, CA. The roundabout, shown in Fig.~\ref{fig:full_structure}(b), is connected with 8 branches and each of them has one entry lane and one exit lane.

The data was collected by a drone from bird’s eye view. To smoothen the noisy data, we performed a downsampling to decrease the sampling rate from 10Hz to 5Hz and applied an Extended Kalman Filter (EKF). We manually picked 1534 driving segments from the collected data, where 80\% were randomly selected for training and 20\% for testing.

\subsubsection{Reference Path}
According to the dataset, a total of 19 reference paths were considered, which can be seen in Fig.~\ref{fig:full_structure}(b). Each reference path corresponds to a routing information from one lane to another and is generated by finding the best fitted path among all vehicle trajectories in the data that have the same entry and exit lane. 

%These reference path are mainly used for intention prediction purpose and can be also utilized for converting the vehicle state into the Frenet Frame. 

%\begin{figure}[htbp]
%	\centering
%	%\epsfig{figure=structure,height=3.8cm}
%	\includegraphics[scale=0.65]{ref_path.pdf}
%	\caption{Roundabout map and all reference path.}
%	\label{fig:ref_path}
%\end{figure}

\begin{figure*}[htbp]
	\centering
	\includegraphics[scale=0.4]{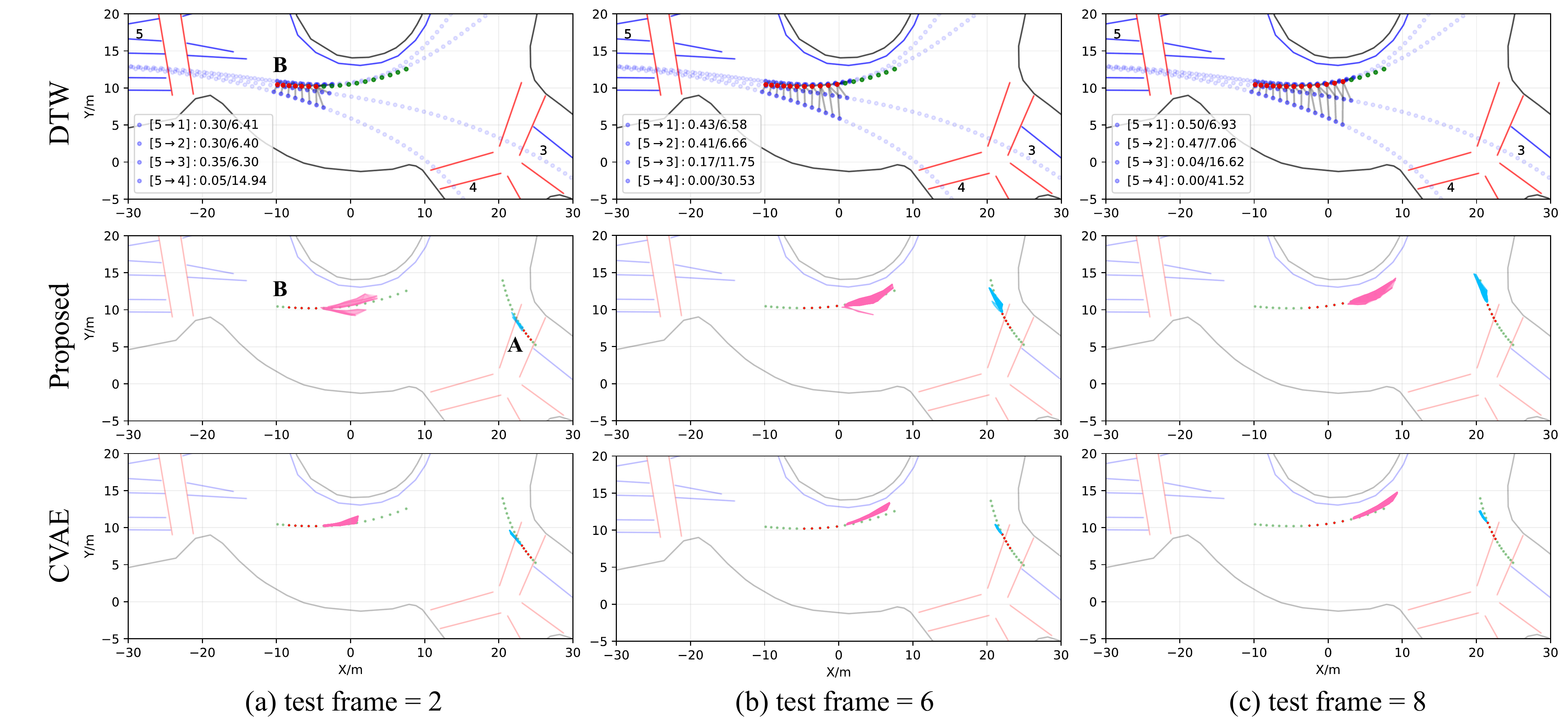}
	\caption{Visualization of the prediction results under a selected test scenario. Each dotted point represents the position of the vehicle. The green dots represent the ground truth trajectory of the vehicle, where the time step between each two dots is 0.2s; the red dots represent the input historical trajectory. Ten future trajectories for car $A$ and car $B$ are sampled from the trained model and are shown in cyan and pink color, respectively. The legend in the plots of the first row should be interpreted as: \textit{[entry lane $\rightarrow$ exit lane]: posterior intention probability / current DTW value}.}
	\label{fig:visualize_traj}
\end{figure*}
\subsection{Problem Description and Feature Selection}
\subsubsection{Problem Description}
For roundabout scenario, one of the most typical interaction scenarios happens when one vehicle (car $A$) is waiting behind the stop/yield sign and trying to enter the circular roadway, while another vehicle (car $B$) is driving on the circular roadway towards the direction of car $A$. Under such circumstance, the potential exit lane for car $B$ will largely influence the driving behavior of both vehicles. For example, if car $B$ exits the circular roadway before reaching the current lane of car $A$, the driving trajectories of two vehicles will not be influenced by each other; contrarily, if car $B$ keep driving on the circular roadway, two cars will begin negotiating with each other to decide who should go first, which will affect their future trajectories. 

As most of the selected cases in our dataset belong to the aforementioned situation, we only consider the driving intent of car $B$ as the intention input $\mathcal{I}$ in our proposed prediction model. Moreover, the front vehicles of car $A$ and car $B$ are regarded as environment information in each driving case, which are essential influence factors of vehicle behaviors.

\subsubsection{Feature Selection}
The input of past joint trajectories contains four features at each time step: $\bm{\xi}^{A,B}_{(t-T1):t} = [\bm{x}^A_{(t-T1):t}, \bm{y}^A_{(t-T1):t}, \bm{x}^B_{(t-T1):t}, \bm{y}^B_{(t-T1):t}]$. The environment input contains the surrounding vehicles' information at the current time step $t$, which is the state of each interactive car's front vehicle: $\mathcal{E} = [\bm{x}^{front}_t, \bm{y}^{front}_t, \bm{v}^{front}_t]$. Here, $x$ and $y$ represent the vehicle's location in Euclidean coordinates, while $v$ denotes the vehicle speed. The driving intention is converted to an one-hot vector, which denotes the intended exit branch for the selected vehicle out of all 8 possible branches.

\subsection{Implementation Details}
As shown in Fig.~\ref{fig:full_structure}, the environment information passes through a fully connected network with 16 neurons and the sequence of past joint trajectories are fed into one LSTM cell with 16 neurons. Both the encoder and decoder contain three fully connected layers of 64 neurons with $tanh$ as non-linear activation function. The latent space dimension is set to 2 and a randomly sampled $z$-vector from a unit Gaussian is used as one of the input of the decoder. In this problem, $T1$ and $T2$ are both set to 5, where we want to predict 1s into the future using the past 1s information. According to the experiment, the method has the best performance when $\beta$ is set to 0.005. 

\begin{figure*}[htbp]
	\centering
	\includegraphics[scale=0.37]{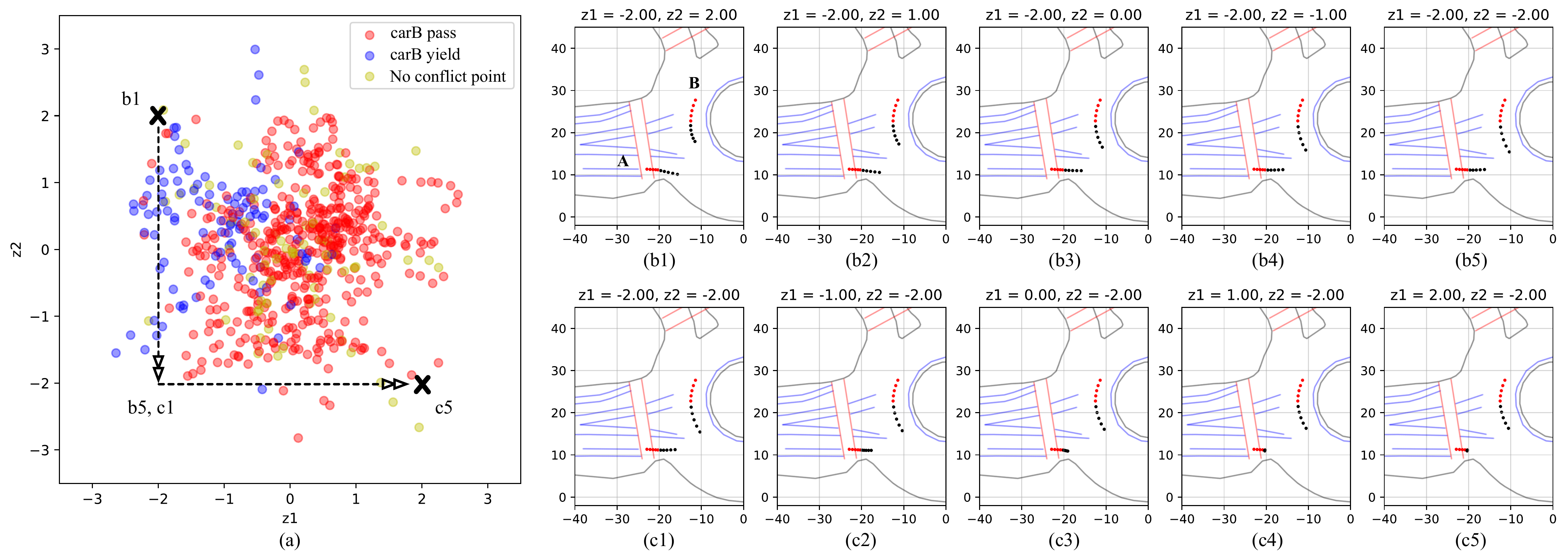}
	\caption{Latent space visualization and demonstration of the interpretable model. In (b1-b5) and (c1-c5), red points represent historical trajectories and black points represent predicted future trajectories for both vehicles.}
	\label{fig:latent_space}
\end{figure*}
\section{Results and Evaluations}
In this section, we first visually illustrate the result of the proposed model through several selected cases. Then we introduce the quantitative evaluation metric and present the comparison result with other baseline methods.

\subsection{Visual Illustration}
\subsubsection{Intention Prediction}
We selected a representative case to demonstrate how intention is predicted using Bayesian update and DTW, shown in the first row of Fig.~\ref{fig:visualize_traj}. The top four reference paths that have the highest intention probability were plotted in light blue and we further selected a path segment (dark blue) from each reference path to calculate the DTW value with the observed historical trajectory (red).

At the beginning, the vehicle's intention is ambiguous and it has similar probability of exiting at branch 1, 2, and 3. However, the reference path of exiting at the 4th branch has the largest DTW value compared to the vehicle's observed trajectory and thus its corresponding intention probability is the smallest among the listed four reference path. As the vehicle continues to move, it drives closer to the roundabout center and has lesser intent to exit. Such behavior is well captured by the intention prediction module as shown in the first and second column of the first row in Fig~\ref{fig:visualize_traj}, where the probabilities of the vehicle exiting the 1st and 2nd branch increase while the likelihood of exiting at the 3rd branch decreases.

\subsubsection{Trajectory Prediction}
We tested and compared the prediction results of our proposed method with the original CVAE approach that does not contain intention inference module. The testing results on a selected scene are shown in the bottom two rows of Fig.~\ref{fig:visualize_traj}. To make a fair comparison, we fixed the 10 randomly sampled latent $z$-values in both methods and used them to generate 10 future joint trajectories of two interacting vehicles. 

According to the result, our proposed method successfully generates different motion patterns which are consistent with the intention prediction result. In contrast, the traditional CVAE method only predicts one motion pattern and it fails to consider the possibility that car $B$ might exit the roundabout at the 3rd branch.

We argue that although two vehicles have interaction and it may influence their future trajectories, the intention of which road to exit for car $B$ is not influenced by the other vehicle, $A$. Therefore, if we are about to predict the joint trajectories of two cars using learning-based methods like CVAE, each vehicle's trajectory will be largely influenced by the data distribution and will not reveal multi-modal property if the amount of data we used are not large enough to cover every possible cases. Even if we have sufficient data, the CVAE network will most likely learn how to closely relate the future motions of two vehicles instead of learning to infer the future joint trajectories based on the historical trajectory of each individual vehicle. In other word, we don't want the network to only learn the motion correlations between two vehicles without treating each of them individually. Hence, the intention factor we added in the proposed method will mitigate such problem by encouraging the network to relate each vehicle's intention to its own past trajectory and then generating its future motions while taking other vehicle's historical motions into consideration.

\subsubsection{Interpretability}
The learned latent space is plotted in Fig.~\ref{fig:latent_space}(a) where we assigned different colors to different interact cases. Although the pass/yield information of two interacting vehicles is not used during training, our proposed method successfully distinguished such motion patterns in the latent space. To illustrate the influence of the sampled $z$-vector to the predicted trajectories, we fixed the intention input $c$ and only changed the $z$ value along its two dimensions. 

As we fix $z1$ and decrease $z2$ (figure b1 to b5), car $B$ increases its speed and shifts from yielding car $A$ to passing car $A$ while the speed of $A$ does not change much. As we fix $z2$ and increase $z1$ (figure c1 to c5), car $B$ always passes car $A$ but the speed of car $A$ gradually decreases. Therefore we can conclude that if we move $z$ from the 2nd to the 4th quadrant of the 2D unit Gaussian, there will be an obvious change of interaction patterns between two cars. Hence, the proposed method is interpretable as the sampled output can be well-explained by the location of $z$. Moreover, being able to generate various motion patterns from different sampled $z$ values can be also regarded as having the multi-modal property.

\subsection{Metric}
\subsubsection{Mean Squared Error (MSE)}
MSE is commonly used to evaluate the prediction performance and the equation can be written as:
%For models generating multi-modal predictive distributions, we use the mode with the highest probability for calculating the MSE. 
\begin{equation}
	%RMSE = \sqrt{\frac{1}{N_s}\sum_{s=1}^{N_s}\big(\bm{\xi}_{t:(t+T_2)} - \hat{\bm{\xi}}_{t:(t+T_2)}^s\big)^2},
	MSE = \frac{1}{N_s}\sum_{s=1}^{N_s}\big(y - \hat{y}^s)^2,
\end{equation}
where $\hat{y}^s$ is the $s$-th sampled prediction result out of $N_s$ output samples and $y$ is the ground-truth.

MSE is skewed in favor of models that average different output modes. In particular, this average may not represent a good prediction when more than one mode exists.
\subsubsection{Negative Log Likelihood (NLL)}
While MSE provides a tangible measure for the predictive accuracy of models, it has limitations while evaluating multi-modal predictions. NLL, instead, is a proper metric for evaluating predictive uncertainty \cite{evaluate_uncertainty} and it can be calculated as:
\begin{equation}
	NLL = \frac{\log \sigma^2(\hat{\bm{y}})}{2} + \frac{(y-\mu(\hat{\bm{y}}))^2}{2\sigma^2(\hat{\bm{y}}))},
\end{equation}
where $\mu(\hat{\bm{y}})$ and $\sigma^2(\hat{\bm{y}})$ represent the mean and variance of $N_s$ output samples respectively.

%What we want to see is that, when the intention is uncertain, the NLL should be small. However, when the intention is very obvious, we expect a 

%\subsubsection{Conditional Log Likelihood (CLL)}
%
%\begin{equation}
%	CLL = 
%\end{equation}

%\subsubsection{Brier Score}

\subsection{Quantitative Performance Evaluation}
We compared our method with the following three approaches, where all of them take historical trajectories as input and generate a sequence of future trajectories as output.
%\subsubsection{Constant Velocity (CV)}
%The vehicle's velocity is assumed to be constant throughout the prediction horizon.
\begin{itemize}
\item \textit{Conditional Variational Autoencoder (CVAE w/o $\mathcal I$)}:
We compared the proposed method with the traditional CVAE approach without using intention prediction module.
\item \textit{Multi-layer Perceptron Ensemble (MLP-ensemble)}:
The MLP is designed to have the same network structure as the decoder in our proposed model. To incorporate uncertainty, we applied the \textit{bagging} strategy to combine predictions of models built on subsets created by bootstrapping.
\item \textit{Monte Carlo dropout (MC-dropout)}:
We also implemented the MC-dropout method \cite{dropout} to estimate the prediction uncertainty by using \textit{Dropout} during training and test time. The mean and variance can be obtained by performing stochastic forward passes and averaging over the outputs.
\end{itemize}

The MSE and NLL values are calculated for all four methods and the results are shown in Table~\ref{tab:comparison}. It is apparent from the table that our  method has the largest value in terms of the MSE but has the smallest NLL. Such results indicate that the proposed method is able to generate output trajectories with the largest uncertainties due to its multi-modal prediction results at the expense of slightly larger MSE value. The reason that other methods have smaller MSE value is because they can only approximate single motion model and all the corresponding output samples are distributed around the groudtruth. However, small MSE is not favorable when the output is supposed to have multiple motion models.

Moreover, we notice that CVAE has comparable results against ML dropout and MLP ensemble in terms of the two evaluation metrics. Therefore, the proposed method, based on the CVAE methodology, is not only able to generate desirable performances, but also capable of estimating explainable multi-modal distributions.

%We want to show that all these models cannot predict multi-modal distribution (create a visualization comparison) and CVAE has comparable performance (compare RMSE, NLL, etc.). Although CVAE may not be the best method, its interpretability of the latent space makes it the best among all these models. Therefore, we further modify the CVAE structure to make it has a multi-modal property. 

\begin{table}[ht]
	\caption{Performance Comparison}
	\label{tab:comparison}
	\centering
	\begin{tabular}{p{1cm} p{1.4cm} p{1.4cm} p{1.4cm} p{1.4cm}}
		\toprule %\toprule[0.1mm]
		Method & Proposed & CVAE Without $\mathcal I$ & MC Dropout & MLP Ensemble \\
		\midrule \midrule
		MSE & 0.45 $\pm$ 0.11 & 0.16 $\pm$ 0.14 & \textbf{0.15 $\pm$ 0.10} & 0.20 $\pm$ 0.11 \\
		\midrule
		NLL & \textbf{0.83 $\pm$ 0.26} & 2.57 $\pm$ 1.02 & 2.10 $\pm$ 0.63 & 3.05 $\pm$ 0.97 \\
		\bottomrule
	\end{tabular}
\end{table}
\section{Conclusions}
In this paper, a multi-modal probabilistic prediction method is proposed, which can predict interactive behavior for traffic participants and acquires interpretability. An exemplar roundabout scenarios with real-world data collected by ourselves was used to demonstrate the performance of our method. First, the prediction results for intention and motion of selected vehicles are visually illustrated through a representative driving case. Then, we plotted the learned latent space to demonstrate the interpretability. Finally, we quantitatively compared the proposed method with three different models: CVAE, MLP ensemble and MC dropout. Our method outperforms these methods in terms of the negative log likelihood metric, which shows its advantages for learning conditional models on multi-modal distributions. In future works, we will focus on making the prediction algorithm not only interpretable but also safe to use under various circumstances.

\section{Acknowledgment}
We thank Junming Chen and Di Wang for data processing works.

\addtolength{\textheight}{-10cm}   % This command serves to balance the column lengths

%%%%%%%%%%%%%%%%%%%%%%%%%%%%%%%%%%%%%%%%%%%%%%%%%%%%%%%%%%%%%%%%%%%%%%%%%%%%%%%%

\bibliographystyle{IEEEtran}
\bibliography{IV2019_ref}

\end{document}